\documentclass[runningheads]{llncs}

\usepackage{eccv}

\usepackage{eccvabbrv}
\usepackage{graphicx}
\usepackage{booktabs}
\usepackage{amsmath}
\usepackage{amssymb}
\usepackage{adjustbox}
\usepackage{multirow}
\usepackage{iftex}
\ifPDFTeX
\usepackage[accsupp]{axessibility}
\fi

\newcommand{\best}[1]{\textbf{#1}}
\newcommand{\second}[1]{#1}

\usepackage[pagebackref,colorlinks=true,linkcolor=blue,citecolor=blue,urlcolor=blue]{hyperref}
\usepackage[capitalize]{cleveref}
\crefname{section}{Sec.}{Secs.}
\Crefname{section}{Section}{Sections}
\Crefname{table}{Table}{Tables}
\crefname{table}{Tab.}{Tabs.}

\begin{document}

\title{UHD Image Deblurring via Autoregressive Flow with Ill-conditioned Constraints}
\titlerunning{Autoregressive Flow for UHD Deblurring}

\author{
Yucheng Xin$^1$ \and
Dawei Zhao$^2$ \and Xiang Chen$^3$\and Chen Wu$^4$\and Pu Wang$^5$\and Dianjie Lu$^1$\and Guijuan Zhang$^1$\and Xiuyi Jia$^3$\and
Zhuoran Zheng$^2$\thanks{Corresponding author.}
}

\authorrunning{Xin et al.}

\institute{
Shandong Normal University
\and
Qilu University of Technology\\
\email{zhengzr@njust.edu.cn}
\and 
Nanjing University of Science and Technology \and 
National University of Defense Technology
\and 
Shandong University
}

\maketitle
\pagestyle{headings}

\begin{abstract}
Ultra-high-definition (UHD) image deblurring poses significant challenges for UHD restoration methods, which must balance fine-grained detail recovery and practical inference efficiency. 
%
Although prominent discriminative and generative methods have achieved remarkable results, a trade-off persists between computational cost and the ability to generate fine-grained detail for UHD image deblurring tasks.
%
To further alleviate these issues, we propose a novel autoregressive flow method for UHD image deblurring with an ill-conditioned constraint.
Our core idea is to decompose UHD restoration into a progressive coarse-to-fine process: at each scale, the sharp estimate is formed by upsampling the previous-scale result and adding a current-scale residual, enabling stable stage-wise refinement from low to high resolutions. 
We further introduce Flow Matching to model residual generation as a conditional vector field and perform few-step ODE sampling with efficient Euler/Heun solvers, enriching details while keeping inference affordable. 
Since multi-step generation at UHD can be numerically unstable, we propose an ill-conditioning suppression scheme by imposing condition-number regularization on a feature-induced attention matrix, improving convergence and cross-scale consistency.
%
Our method demonstrates promising performance on blurred images at 4K (3840$\times$2160) or higher resolutions.
\end{abstract}

\section{Introduction}

Image blurring is prevalent in scenarios such as handheld shooting, moving objects, low light, and long exposure, significantly degrading visual perception for human observers and causing adverse effects on downstream vision tasks like detection, recognition, and reconstruction. 
With the increasing adoption of ultra-high-definition (UHD) imaging in mobile devices and professional imaging equipment, deblurring methods are facing a more acute dilemma: the exponential increase in resolution brings richer textures and structural details, but also pushes the requirements for computation, memory, and latency to their limits. 
%

\begin{figure}[t]
    \centering
    \includegraphics[width=\columnwidth]{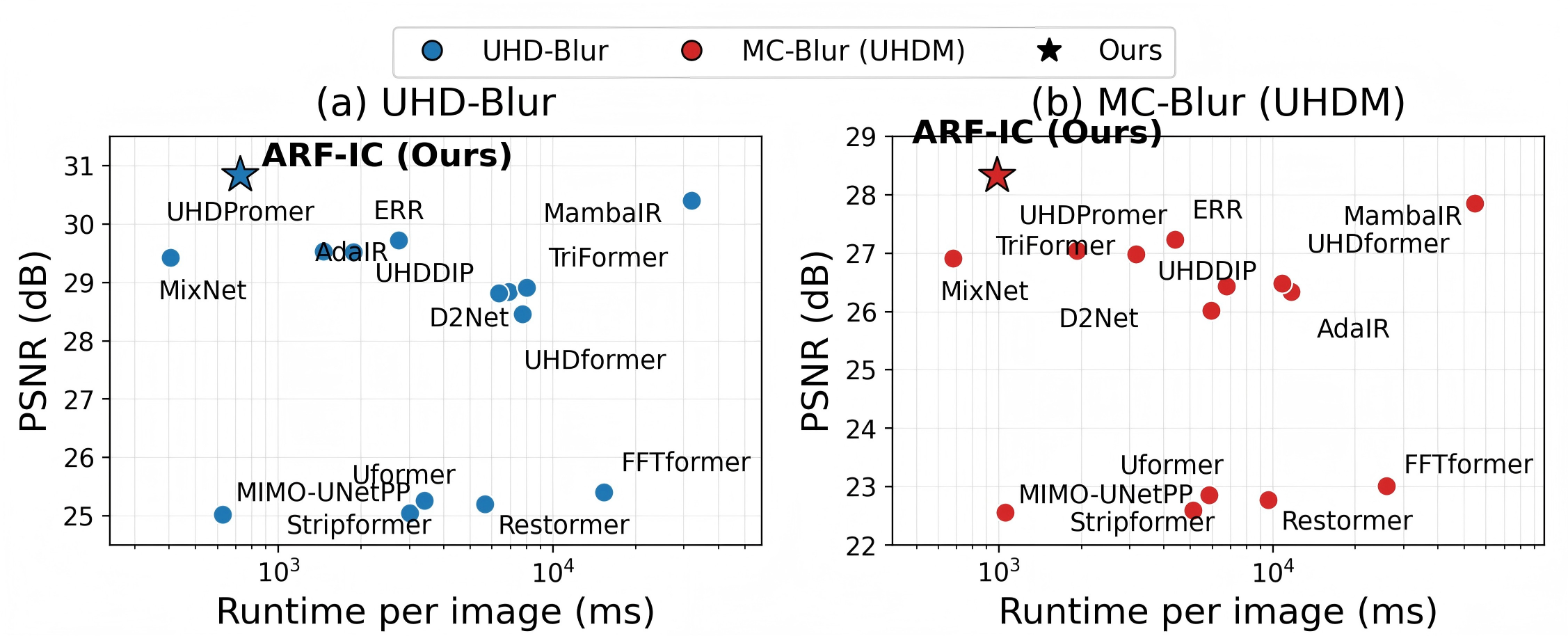}
    \caption{Runtime vs. Quality trade-off on UHD deblurring benchmarks. We plot PSNR against per-image runtime (log-scale, in milliseconds) for representative baselines and ARF-IC (Ours) on UHD-Blur and MC-Blur (UHDM).}
    \vspace{-6mm}
    \label{fig:runtime_psnr_tradeoff}
\end{figure}

Recent Transformer-based architectures~\cite{vaswani2017attention} have been used to model longer-range dependencies, pushing the state-of-the-art performance on several benchmarks. However, these methods often rely on deeply stacked MLPs or computationally expensive global attention units, easily encountering the bottleneck of ``manageable training but slow inference, or feasible inference at prohibitively high cost'' in UHD imaging scenarios. 
Meanwhile, generative modeling approaches such as diffusion models (e.g., DDPM~\cite{ho2020ddpm}) and score-based SDE models~\cite{song2021score} provide stronger detail reconstruction capabilities, but their iterative sampling becomes even less tolerable at UHD. Fast samplers such as DDIM~\cite{song2020ddim} and DPM-Solver~\cite{lu2022dpmsolver} reduce the step budget, yet when the model must ``invent'' missing high-frequency details, small numerical or modeling errors can be amplified into visually salient artifacts, resulting in unstable textures or hallucinated patterns; Fig.~\ref{fig:runtime_psnr_tradeoff} visualizes this runtime--PSNR trade-off on UHD-Blur (4K resolution) and MC-Blur UHDM ($\geq$ 4K resolution), clarifying why strong UHD restoration quality can quickly become computationally expensive and motivating a framework that is both high-quality and practically fast.

This paper focuses on the stable generation and efficiency issues in UHD deblurring, proposing an autoregressive flow with ill-conditioned constraints suitable for deployment on consumer-grade GPUs (RTX 3090).
Specifically, we explicitly define the deblurring process as a generative pipeline that progresses from coarse to fine: starting from low-resolution images, it progressively samples toward high-resolution images. Unlike discrete iterative diffusion sampling, we learn conditional vector fields based on flow matching~\cite{lipman2022flowmatching} (rectified flow~\cite{liu2022rectifiedflow}) and model residual generation as a continuous-time dynamic evolution process.
%
During inference, efficient ODE numerical solvers (e.g., Euler or Heun) are employed to complete the integration from noise to residual with a small number of steps, significantly reducing UHD inference overhead while maintaining generative capability. To enhance cross-scale generalization and reuse, we introduce a joint time-scale embedding, enabling the same vector field network to operate stably under different integration timesteps and scale conditions.

Compared to traditional multi-scale regression networks or autoregressive models, our approach models each refinement process from low to high resolution as a continuous spatial model.
%
Notably, such generative models are highly susceptible to disruption from data noise, the discretization of ordinary differential equations, and the accumulation of cross-scale errors, which can hinder model convergence. Therefore, we characterize this instability through the condition number of the induced interaction matrix via feature mixing and propose a condition number regularization method to limit amplification effects in worst-case scenarios, thereby stabilizing the generation process from coarse to fine residuals. 
Our approach was evaluated on the embedded side, demonstrating stable real-time operation on consumer-grade mobile devices (rendering a 4K image in under 2 seconds, Apple or Huawei phones).
Our main \textbf{contributions} are summarized as follows:

\vspace{-2mm}
\begin{enumerate}
    \item We propose an Autoregressive Flow method tailored for UHD image deblurring, which decomposes full-resolution restoration into a coarse-to-fine, scale-by-scale process of residual generation and fusion, thereby achieving scalable high-resolution deblurring.
    \item We deepen the stability analysis of UHD generative restoration from the perspectives of error amplification and numerical conditioning, and organically integrate condition-number regularization, a resolution-aware pre-inference downsampling strategy, and an analytical detail compensation term to enable efficient texture-preserving inference for UHD images.
    \item Extensive experiments are conducted on two UHD deblurring datasets and multiple standard-resolution deblurring benchmarks. The results demonstrate that the proposed method outperforms current state-of-the-art approaches in terms of both accuracy and speed, validating its effectiveness and scalability in UHD deployment.
\end{enumerate}

\section{Related Work}

\noindent\textbf{UHD Image Deblurring.}
UHD deblurring exposes a deployment gap between strong restoration quality and practical inference constraints (memory, latency, and global consistency). Existing pipelines typically rely on multi-scale designs, patch-wise/tiling inference, or low-resolution processing followed by upsampling and refinement. While effective, tiling can introduce seams and long-range inconsistency, and aggressive downsampling can lose fine textures. Recent UHD restoration backbones include UHDformer~\cite{wang2024uhdformer}, MixNet~\cite{wu2024mixnet}, TriFormer~\cite{ma2025triformer}, AdaIR~\cite{cui2024adair}, and D2Net~\cite{wu2024d2net}, and more recently UHDPromer~\cite{wang2026uhdpromer} further improves efficiency for UHD restoration including deblurring. Benchmark and solution works such as UHDDIP~\cite{wang2024uhddip} and ERR~\cite{zhao2025err} further highlight UHD-specific degradations and efficiency constraints. 
Although they outperform in visual perception, they are difficult to deploy on consumer-grade GPUs.

Our approach is a low-cost generative UHD deblurring algorithm (autoregressive method~\cite{liu2025conditional,guo2022lar,qu2025visual,kong2025nsarm,wei2025perceive,rajagopalan2025restorevarvisualautoregressivegeneration}) that utilizes ill-posed constraints to maintain the stability of generated details.

\begin{figure*}[t]
    \centering
    \includegraphics[width=\textwidth]{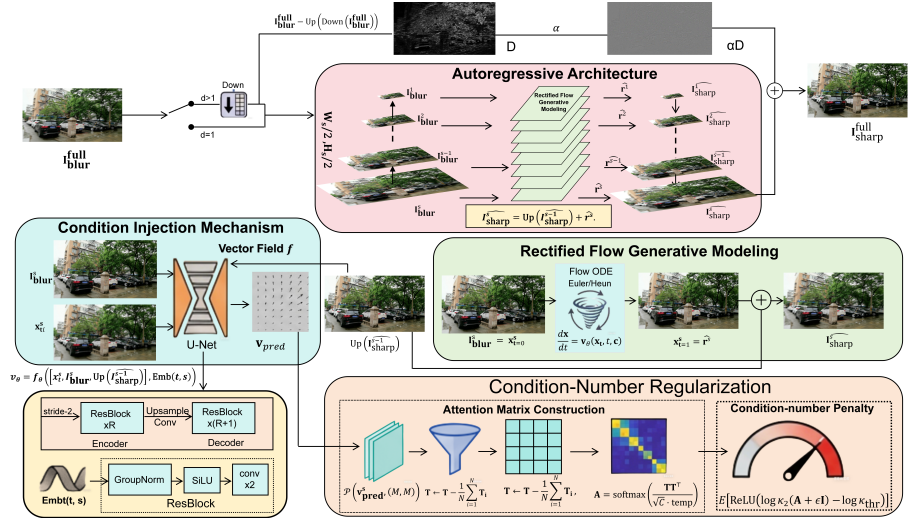}
    \caption{The overall framework of the proposed UHD deblurring method. We begin with coarse-to-fine reconstruction using rectified residual sampling, stabilizing the model's regression capability by constraining the ill-conditionedness of features.}
    \vspace{-6mm}
    \label{fig:method_framework}
\end{figure*}

\section{Methodology}

\subsection{Overview}

Given a blurred image $\mathbf{I}_{\text{blur}}$, we propose an Autoregressive Flow method for efficient and stable UHD deblurring. The restoration progresses from coarse to fine, where the estimate at scale $s$ is formed by fusing an upsampled previous prediction with a scale-specific residual: $\hat{\mathbf{I}}_{\text{sharp}}^{s} = \text{Up}(\hat{\mathbf{I}}_{\text{sharp}}^{s-1}) + \hat{\mathbf{r}}^{s}$. The residual $\hat{\mathbf{r}}^{s}$ is generated by solving a short ODE driven by a conditional vector field learned via flow matching, enabling few-step Euler/Heun sampling. 
The overall framework is shown in Fig.~\ref{fig:method_framework}.

\subsection{Autoregressive Architecture}
%
Concretely, we index scales by $s\in\{1,\dots,S\}$ from coarse to fine, and denote by $\mathbf{I}_{\text{blur}}^s$ and $\mathbf{I}_{\text{sharp}}^s$ the blurry input and the corresponding sharp target resized to $(H_s,W_s)$. This multi-scale scheme also clarifies the role of the previous-scale prediction: it serves as a strong, low-frequency prior that guides the generation at the current scale, rather than requiring the model to ``re-create'' the entire image from scratch at each resolution.

\subsubsection{Scale Sequence Construction}

We construct the scale sequence $(H_s, W_s)$ for $s=1,\ldots,S$ by recursively downsampling from the target resolution $(H, W)$. At each downsampling step, the height and width are divided by 2 with ceiling rounding, i.e., $H_{s-1} = \lceil H_s / 2 \rceil$ and $W_{s-1} = \lceil W_s / 2 \rceil$, until the shortest side $\min(H_1, W_1) \leq \tau_{\min}=4$ pixels. This yields a scale sequence arranged from small to large, where $s=1$ corresponds to the coarsest resolution and $s=S$ corresponds to the target resolution. Intuitively, the coarsest scales capture the scene layout and dominant edges with low computational burden, while the finer scales focus on refining textures and small structures under the guidance of the coarse prior.

\subsubsection{Autoregressive Residual Fusion}

At each scale $s$, we autoregressively fuse the result from the previous scale through residual addition:
\begin{equation}
\hat{\mathbf{I}}_{\text{sharp}}^s = \text{Up}(\hat{\mathbf{I}}_{\text{sharp}}^{s-1}) + \hat{\mathbf{r}}^s,
\end{equation}
where $\text{Up}(\cdot)$ denotes bilinear upsampling to the spatial dimensions $(H_s, W_s)$ of the current scale, $\hat{\mathbf{I}}_{\text{sharp}}^{s-1}$ is the sharp estimate from the previous scale, and $\hat{\mathbf{r}}^s$ is the residual predicted at the current scale. This design reduces the modeling burden: instead of regenerating the full image, the model only needs to explain what is newly resolvable at the current resolution. It also tends to make the target distribution at each scale more concentrated (residuals are typically smaller in magnitude), which is favorable for stable few-step sampling.
During training, the target residual is defined as:
\begin{equation}
\mathbf{r}^s = \mathbf{I}_{\text{sharp}}^s - \text{Up}(\mathbf{I}_{\text{sharp}}^{s-1}),
\end{equation}
where $\mathbf{I}_{\text{sharp}}^s$ and $\mathbf{I}_{\text{sharp}}^{s-1}$ are the ground-truth sharp images at the current and previous scales, respectively. For the base case $s=1$, we define $\mathbf{I}_{\text{sharp}}^{0}=\mathbf{0}$ (and thus $\text{Up}(\mathbf{I}_{\text{sharp}}^{0})=\mathbf{0}$), so $\mathbf{r}^{1}=\mathbf{I}_{\text{sharp}}^{1}$.

We adopt a teacher forcing strategy, using the ground-truth sharp image from the previous scale as the prior (rather than the predicted value) to avoid error accumulation affecting training stability. During inference, the prior is replaced by the predicted $\hat{\mathbf{I}}_{\text{sharp}}^{s-1}$. In this setting, the residual formulation (predicting only the scale-wise increment) and the conditioning on $\mathbf{I}_{\text{blur}}^s$ help mitigate cross-scale error amplification, and an optional consistency term (Sec.~Loss Function) can further stabilize the trajectory.

\subsubsection{Condition Injection Mechanism}

The current-scale generation should be aware of (i) the partially generated content from the previous scale and (ii) the observation at the current scale. We therefore integrate multi-source information through channel concatenation:
\begin{equation}
\mathbf{v}_\theta = f_\theta\left( [\mathbf{x}_t^s, \mathbf{I}_{\text{blur}}^s, \text{Up}(\hat{\mathbf{I}}_{\text{sharp}}^{s-1})], \text{Emb}(t, s) \right),
\end{equation}
where the square brackets $[\cdot]$ denote channel-wise concatenation. The three input components are: the intermediate state at the current time $\mathbf{x}_t^s$ (3 channels), the blurred image at the current scale $\mathbf{I}_{\text{blur}}^s$ (3 channels), and the upsampled sharp estimate from the previous scale $\text{Up}(\hat{\mathbf{I}}_{\text{sharp}}^{s-1})$ (3 channels), totaling 9 channels. For the coarsest scale ($s=1$), we set $\text{Up}(\hat{\mathbf{I}}_{\text{sharp}}^{0})=\mathbf{0}$.

$\text{Emb}(t, s)$ is the joint embedding of time and scale, which is added to intermediate features through linear projection in each residual block, achieving global modulation of spatiotemporal conditions. The scale condition scalar is normalized as $s_{\text{scalar}} = (s-1)/(S-1) \in [0,1]$, where $S$ is the total number of scales, ensuring scale information is uniformly distributed in the $[0,1]$ interval. 

\subsection{Rectified Flow Generative Modeling}

With the autoregressive prior $\text{Up}(\hat{\mathbf{I}}_{\text{sharp}}^{s-1})$ providing a coarse estimate, the remaining task at scale $s$ becomes generating the residual $\hat{\mathbf{r}}^s$ that completes the image. We model this residual distribution with Rectified Flow, so that sampling reduces to solving a short ODE whose dynamics are conditioned on the blurry observation and the previous-scale prior. 

\subsubsection{Rectified Flow Basics}

Rectified Flow~\cite{liu2022rectifiedflow} is a class of generative models based on Continuous Normalizing Flow~\cite{grathwohl2019ffjord} and Neural ODE~\cite{chen2018neuralode} formulations that map a simple prior distribution (e.g., Gaussian noise) to the target data distribution by learning a time-varying vector field. Its core is solving an ordinary differential equation (ODE):
\begin{equation}
\frac{d\mathbf{x}}{dt} = \mathbf{v}_\theta(\mathbf{x}_t, t, \mathbf{c}), \quad \mathbf{x}(0) = \mathbf{z}_0 \sim \mathcal{N}(0, \sigma_0^2 \mathbf{I}),
\end{equation}
where $\mathbf{x}_t$ denotes the state variable at time $t$, $\mathbf{v}_\theta$ is the parameterized vector field network, $\mathbf{c}$ is the conditioning information (in this work, including the blurred image and the result from the previous scale), $\mathbf{z}_0$ is the initial noise, $\sigma_0$ is the noise standard deviation (default 1.0), and $\mathbf{I}$ is the identity matrix. This equation describes the continuous evolution trajectory from the noise distribution to the target distribution.

\subsubsection{Flow Matching}

Given the target residual $\mathbf{r}^s$ at scale $s$, we construct intermediate states using a linear interpolation path in Flow Matching~\cite{lipman2022flowmatching}:
\begin{equation}
\mathbf{x}_t^s = (1-t) \mathbf{z}_0 + t \mathbf{r}^s, \quad t \sim \mathcal{U}(0,1),
\end{equation}
where $t$ is randomly sampled from the uniform distribution $\mathcal{U}(0,1)$. This simple path yields a constant target vector field,
\begin{equation}
\mathbf{v}_{\text{target}}^s = \mathbf{r}^s - \mathbf{z}_0,
\end{equation}
which makes training particularly stable: the network only needs to predict the straight-line transport direction under the given condition $\mathbf{c}^s$. The training loss is the mean squared error:
\begin{equation}
\mathcal{L}_{\text{flow}} = \frac{1}{S} \sum_{s=1}^S \mathbb{E}_{t, \mathbf{z}_0} \left[ \left\| \mathbf{v}_\theta(\mathbf{x}_t^s, t, \mathbf{c}^s) - \mathbf{v}_{\text{target}}^s \right\|_2^2 \right],
\end{equation}
where $S$ is the total number of scales. 

\subsubsection{ODE Solving}

During inference, sampling $\hat{\mathbf{r}}^s$ corresponds to integrating the ODE from $t=0$ to $t=1$ starting from noise. Since each scale only needs to produce a residual (rather than a full image), we can use very few integration steps without sacrificing quality. We employ the Heun method (second-order Runge-Kutta) for numerical integration. Let the step size be $\Delta t = 1/N_s$, where $N_s$ is the number of sampling steps at the current scale. The update rule is:
\begin{align}
\mathbf{x}_e &= \mathbf{x}_k + \Delta t \cdot \mathbf{v}_\theta(\mathbf{x}_k, t_k, \mathbf{c}^s), \\
\mathbf{x}_{k+1} &= \mathbf{x}_k + \frac{\Delta t}{2} \left[ \mathbf{v}_\theta(\mathbf{x}_k, t_k, \mathbf{c}^s) + \mathbf{v}_\theta(\mathbf{x}_e, t_{k+1}, \mathbf{c}^s) \right],
\end{align}
where $t_k = k \Delta t$. The final residual estimate is $\hat{\mathbf{r}}^s = \mathbf{x}(t=1)$. In our implementation, $N_s$ is specified by a coarse-to-fine schedule (default $[8, 8, 6, 4, 3, 2, 1]$). If the actual number of scales $S$ differs, we truncate the schedule to the first $S$ entries (when $S \leq 7$) or pad it by repeating the last value (when $S > 7$). When $N_s=1$, Heun reduces to a single Euler step, and we use the Euler update for efficiency.

In addition, supplementary materials provide further details regarding network structure and time step embedding.

\subsection{Condition-Number Regularization}

\subsubsection{Attention Matrix Construction}

Given the vector field prediction $\mathbf{v}_{\text{pred}}^s \in \mathbb{R}^{B \times 3 \times H_s \times W_s}$ at scale $s$ (where $B$ is the batch size, 3 is the number of RGB channels, and $H_s$ and $W_s$ are the spatial dimensions), we construct an ``induced attention matrix'' to quantify the similarity structure among features:

\noindent i) \textbf{Adaptive Pooling}: Pool the features to a fixed grid:
\begin{equation}
\mathbf{F} = \mathcal{P}\!\left(\mathbf{v}_{\text{pred}}^s, (M, M)\right),
\end{equation}
where $\mathcal{P}(\cdot)$ denotes adaptive average pooling (\texttt{AdaptiveAvgPool2d}), and $M=16$ (default). This operation uniformly downsamples features of arbitrary spatial dimensions to a $16 \times 16$ grid, reducing subsequent computational complexity.

\noindent ii)  \textbf{Tokenization and Normalization}: Flatten to $\mathbf{T} \in \mathbb{R}^{B \times N \times C}$, where $N=M^2=256$ is the number of spatial positions (token count) and $C=3$ is the number of channels. Perform centering and L2 normalization on each token:
\begin{equation}
\mathbf{T} \leftarrow \mathbf{T} - \frac{1}{N}\sum_{i=1}^N \mathbf{T}_i, \qquad
\mathbf{T}_i \leftarrow \frac{\mathbf{T}_i}{\|\mathbf{T}_i\|_2 + \varepsilon},
\end{equation}
where the centering subtracts the mean over the token dimension (for each channel), and the normalization is applied along the channel dimension with $\varepsilon=10^{-6}$. This makes the subsequent similarity computation depend mainly on direction rather than magnitude.

\noindent iii) \textbf{Self-Attention Computation}:
\begin{equation}
\mathbf{A} = \text{softmax}\left( \frac{\mathbf{T} \mathbf{T}^\top}{\sqrt{C}\cdot \text{temp}} \right) \in \mathbb{R}^{B \times N \times N},
\end{equation}
where $\mathbf{T} \mathbf{T}^\top$ computes token-pair similarities, $\sqrt{C}$ is the standard attention scaling, $\text{temp}$ is an optional temperature (default $1$), and $\text{softmax}$ normalizes along the last dimension.

\subsubsection{Condition Number Constraint}

To prevent $\mathbf{A}$ from approaching singularity (maximum singular value much larger than minimum singular value), we add a small diagonal term and penalize large condition numbers:
\begin{equation}
\mathcal{L}_{\text{cond}} = \mathbb{E} \left[ \text{ReLU}\left( \log \kappa_2(\mathbf{A} + \epsilon \mathbf{I}) - \log \kappa_{\text{thr}} \right) \right],
\end{equation}
where $\epsilon=10^{-3}$, $\kappa_2(\cdot)$ is the spectral (2-norm) condition number (implemented via \texttt{torch.linalg.cond}), and $\kappa_{\text{thr}}=100$. In the implementation, we compute this penalty on a small subsample of the batch (default: at most one image) and clamp condition numbers via \texttt{nan\_to\_num} for numerical safety. 
\textbf{Essentially, the condition number reflects a matrix's sensitivity.} This regularization term is applied only to the final model's output scale to save training time. Furthermore, an excessively large condition number indicates that the attention structure is approaching singularity, which amplifies perturbations and compromises training/inference stability.

\subsubsection{Why the ill-conditioned constraint matters ?}
In UHD autoregressive residual generation with few-step ODE sampling, small errors from discretization, upsampling, and feature mixing are inevitable. The key issue is whether these perturbations get amplified. We therefore measure the worst-case amplification capability of the feature-induced attention matrix using the spectral (2-norm) condition number. Define:
\begin{equation}
\tilde{\mathbf A}=\mathbf A+\varepsilon\mathbf I,\qquad
\kappa_2(\tilde{\mathbf A})=\|\tilde{\mathbf A}\|_2\|\tilde{\mathbf A}^{-1}\|_2
=\frac{\sigma_{\max}(\tilde{\mathbf A})}{\sigma_{\min}(\tilde{\mathbf A})},
\end{equation}
where $\kappa_2$ increases sharply when $\sigma_{\min}$ is small (near singularity), meaning the attention structure is ill-conditioned; the diagonal shift $\varepsilon\mathbf I$ prevents exact singularity and stabilizes computation.

To connect this to stability, view attention mixing locally as a linear map $\mathbf y=\tilde{\mathbf A}\mathbf x$. For an input perturbation $\delta\mathbf x$, we have $\delta\mathbf y=\tilde{\mathbf A}\delta\mathbf x$, and the singular-value bounds yield
\begin{equation}
\frac{\|\delta\mathbf y\|_2}{\|\mathbf y\|_2}
=\frac{\|\tilde{\mathbf A}\delta\mathbf x\|_2}{\|\tilde{\mathbf A}\mathbf x\|_2}
\le
\frac{\sigma_{\max}(\tilde{\mathbf A})}{\sigma_{\min}(\tilde{\mathbf A})}
\cdot\frac{\|\delta\mathbf x\|_2}{\|\mathbf x\|_2}
=
\kappa_2(\tilde{\mathbf A})\cdot\frac{\|\delta\mathbf x\|_2}{\|\mathbf x\|_2}.
\end{equation}
This shows $\kappa_2(\tilde{\mathbf A})$ directly upper-bounds worst-case relative error amplification; thus, a large condition number implies that tiny perturbations can be magnified, harming multi-step ODE trajectory consistency and cross-scale fusion stability in UHD.

Motivated by this amplification bound, we penalize only overly ill-conditioned cases to suppress instability while preserving capacity:
\begin{equation}
\mathcal L_{\mathrm{cond}}
=
\mathbb E\Big[\mathrm{ReLU}\big(\log \kappa_2(\mathbf A+\varepsilon\mathbf I)-\log \kappa_{\mathrm{thr}}\big)\Big].
\end{equation}
The $\log(\cdot)$ compresses the dynamic range for stable optimization, and $\mathrm{ReLU}$ activates the penalty only when $\kappa_2$ exceeds $\kappa_{\mathrm{thr}}$. Therefore, the ill-conditioned constraint is a principled, stability-driven regularizer that directly targets worst-case error amplification, making it both reasonable and important.

\subsection{Resolution Control Strategy}





To avoid high-frequency information loss caused by downsampling, we extract detail residuals from the original blurred image and add them back:
\begin{gather}
\mathbf{D} = \mathbf{I}_{\text{blur}}^{\text{full}} - 
\text{Up}(\text{Down}(\mathbf{I}_{\text{blur}}^{\text{full}})), \\
\hat{\mathbf{I}}_{\text{sharp}}^{\text{full}} =
\text{Up}(\hat{\mathbf{I}}_{\text{sharp}}^{\text{low}}) + \alpha \mathbf{D},
\end{gather}
where $\mathbf{I}_{\text{blur}}^{\text{full}}$ is the original full-resolution blurred image, and $\text{Down}(\cdot)$ and $\text{Up}(\cdot)$ are bilinear downsampling/upsampling operations, respectively. The first line computes a Laplacian-pyramid style detail layer. The second line upsamples the low-resolution prediction $\hat{\mathbf{I}}_{\text{sharp}}^{\text{low}}$ and adds the weighted detail $\alpha \mathbf{D}$. In the implementation, this compensation is only applied when the input was actually downsampled (i.e., $d>1$).
The weight $\alpha$ controls the texture--efficiency trade-off (default $\alpha=1.0$); when the input contains strong noise or unreliable high-frequency components, $\alpha$ can be reduced (or the compensation can be disabled) to avoid amplifying artifacts.

\subsection{Loss Function}

The overall objective balances four effects: learning the correct vector field ($\mathcal{L}_{\text{flow}}$), providing a direct end-point supervision signal ($\mathcal{L}_{\text{final}}$), discouraging inconsistent trajectories ($\mathcal{L}_{\text{cons}}$), and suppressing ill-conditioned feature interactions ($\mathcal{L}_{\text{cond}}$).

The total loss consists of four components with configurable weights:
\begin{equation}
\mathcal{L}_{\text{total}} = w_{\text{flow}} \mathcal{L}_{\text{flow}} + w_{\text{final}} \mathcal{L}_{\text{final}} + w_{\text{cons}} \mathcal{L}_{\text{cons}} + w_{\text{cond}} \mathcal{L}_{\text{cond}}.
\end{equation}

Flow matching loss $\mathcal{L}_{\text{flow}}$ is the primary training objective (default $w_{\text{flow}}=1.0$). To accelerate convergence, we additionally introduce a cheap final supervision loss $\mathcal{L}_{\text{final}}$ (default $w_{\text{final}}=1.0$). Concretely, we run a single vector-field forward pass at $t=1$ with a zero residual state and zero previous estimate:
\begin{equation}
\tilde{\mathbf{r}} = \mathbf{v}_\theta\big([\mathbf{0}, \mathbf{I}_{\text{blur}}^{\text{low}}, \mathbf{0}],\, t{=}1,\, s_{\text{scalar}}{=}1\big), \qquad
\tilde{\mathbf{I}}_{\text{sharp}}^{\text{low}} = \tilde{\mathbf{r}},
\end{equation}
then restore it to the original resolution (upsampling plus optional analytical detail compensation) and compute an L1 loss:
\begin{equation}
\mathcal{L}_{\text{final}} = \left\| \text{Restore}(\tilde{\mathbf{I}}_{\text{sharp}}^{\text{low}}) - \mathbf{I}_{\text{sharp}}^{\text{full}} \right\|_1,
\end{equation}
where $\mathbf{I}_{\text{blur}}^{\text{low}}$ is the (optionally downsampled) blurry input used by the model during training, $\text{Restore}(\cdot)$ denotes the restore operation, $\mathbf{I}_{\text{sharp}}^{\text{full}}$ is the ground-truth full-resolution sharp image, and $\|\cdot\|_1$ is the L1 norm.
This auxiliary loss is introduced purely for faster convergence and does not replace ODE-based multi-step sampling at test time; empirically, it provides a direct ``one-pass'' supervision signal that complements flow matching. Note that even when $\mathbf{I}_{\text{blur}}^{\text{low}}$ is obtained by downsampling, we set $s_{\text{scalar}}{=}1$ in this auxiliary pass to align the supervision with the final-scale restoration target after $\text{Restore}(\cdot)$; it is a heuristic training aid rather than a statement about the sampling trajectory.

Consistency constraint $\mathcal{L}_{\text{cons}}$ extrapolates from the current time to the endpoint using the predicted vector field and computes the L1 distance to the target residual to enhance trajectory consistency:
\begin{equation}
\mathcal{L}_{\text{cons}} = \frac{1}{S} \sum_{s=1}^S \mathbb{E} \left[ \left\| \mathbf{x}_t^s + (1-t) \mathbf{v}_{\text{pred}}^s - \mathbf{r}^s \right\|_1 \right],
\end{equation}
where $\mathbf{x}_t^s + (1-t) \mathbf{v}_{\text{pred}}^s$ represents the estimated position at the endpoint by extrapolating from the current state $\mathbf{x}_t^s$ along the predicted vector field $\mathbf{v}_{\text{pred}}^s$ for the remaining time $(1-t)$, and $\mathbf{r}^s$ is the ground-truth target residual. This constraint is enabled by default (default $w_{\text{cons}}=0.1$) and can be disabled by setting $w_{\text{cons}}=0.0$. Condition number regularization $\mathcal{L}_{\text{cond}}$ is enabled by default with $w_{\text{cond}}=0.01$.

\section{Experiments}
\label{sec:experiments}


\subsection{Implementation Details and Datesets}
\label{sec:impl_details}

All experiments are implemented in PyTorch and conducted on a single NVIDIA RTX~3090 GPU. We train the model using paired blurry and sharp images with random cropping and data augmentation. Specifically, we use a crop size of $512\times512$ with random horizontal/vertical flips, and normalize all images to $[-1,1]$ before feeding them into the network. We optimize the network using AdamW~\cite{loshchilov2019adamw} with an initial learning rate of $2\times10^{-4}$ and weight decay $0$. Training is performed for 50 epochs with batch size 8, mixed-precision training (AMP)~\cite{micikevicius2018mixed}, gradient clipping~\cite{pascanu2013difficulty} (max norm 1.0), and an exponential moving average (EMA)~\cite{polyak1992acceleration} of model weights with decay 0.999.

We evaluate on six datasets covering both UHD deblurring and high-resolution (but non-UHD) deblurring.
We first evaluate on two UHD-scale datasets:
UHD-Blur \cite{wang2024uhdformer}, a synthetic motion-blur UHD dataset (approximately $3840\times2160$ and higher) with about 2K paired samples for UHD evaluation; and
MC-Blur (UHDM) \cite{zhang2021mcblur}, a large-scale UHD dataset with synthetic motion blur and large-kernel degradations (tens of thousands of images), which serves as a multi-factor generalization testbed.
We further evaluate on four commonly used high-resolution benchmarks:
GoPro \cite{nah2017gopro}, DVD \cite{su2017dvd}, RealBlur-J and RealBlur-R \cite{rim2020realblur}, which contain blurry and sharp pairs with realistic degradations.

All comparison algorithms employ the same algorithmic environment as ours, including cropped patch resolution, learning rate, learning mechanism (optimizer), and number of iterations.
\subsection{Experimental Results}
\label{sec:comparison}

\subsubsection{UHD Image Deblurring}
\begin{figure}[t]
    \centering
    \includegraphics[width=\columnwidth]{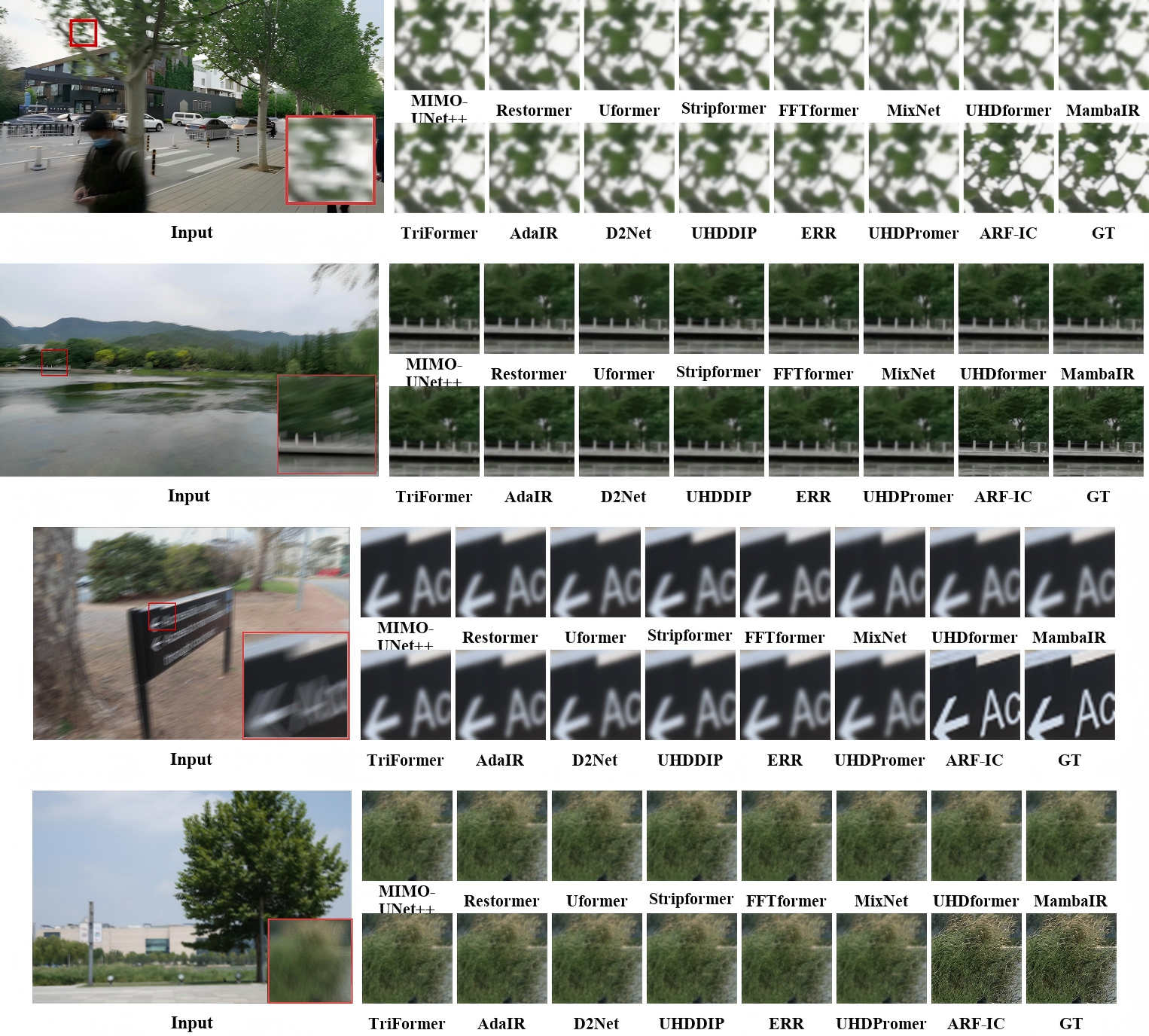}
    \caption{Qualitative comparison on UHD deblurring benchmarks. From top to bottom, the first and second rows are UHD-Blur, and the third and fourth rows are MC-Blur(UHDM).}
    \vspace{-6mm}
    \label{fig:vis_uhd}
\end{figure}

We first evaluate on two UHD deblurring benchmarks, UHD-Blur and the UHDM subset of MC-Blur, and compare with representative recent restoration backbones. Table~\ref{tab:uhd_results} reports parameter counts, computational cost (FLOPs, in G), PSNR, and SSIM~\cite{wang2004ssim}, as well as inference time per-image (s), on both datasets. 
Overall, our approach achieves optimal performance on quality evaluation metrics and demonstrates relatively optimal temporal inference speed.
Qualitative visual comparisons on the two UHD benchmarks are provided in Fig.~\ref{fig:vis_uhd}. 

\begin{table}[t]
\centering
\small
\caption{Quantitative comparison on UHD deblurring benchmarks (PSNR, SSIM, per-image inference time in seconds, and FLOPs in G).}
\label{tab:uhd_results}
\setlength{\tabcolsep}{3.8pt}

\begin{adjustbox}{width=\columnwidth}
\begin{tabular}{@{}ll r r r r r r r r r@{}}
\toprule
Method & Source & {Params$\downarrow$ (M)} & \multicolumn{4}{c}{UHD-Blur} & \multicolumn{4}{c}{MC-Blur(UHDM)} \\
\cline{4-7}\cline{8-11}
& & & {PSNR$\uparrow$} & {SSIM$\uparrow$} & {Time$\downarrow$ (s)} & {FLOPs$\downarrow$ (G)} & {PSNR$\uparrow$} & {SSIM$\uparrow$} & {Time$\downarrow$ (s)} & {FLOPs$\downarrow$ (G)} \\
\midrule
MIMO-UNet++~\cite{cho2021mimo} & ICCV'22 & 16.10 & 25.025 & 0.7517 & \underline{\second{0.625}} & 20,061 & 22.565 & 0.7328 & 1.055 & 27,082 \\
Restormer~\cite{zamir2022restormer} & CVPR'22 & 26.10 & 25.210 & 0.7522 & 5.647 & 49,693 & 22.780 & 0.7341 & 9.611 & 67,086 \\
Uformer~\cite{wang2022uformer} & CVPR'22 & 20.60 & 25.267 & 0.7515 & 3.400 & 27,436 & 22.857 & 0.7329 & 5.846 & 37,038 \\
Stripformer~\cite{tsai2022stripformer} & ECCV'22 & 19.70 & 25.052 & 0.7501 & 2.999 & 49,392 & 22.602 & 0.7307 & 5.097 & 66,679 \\
FFTformer~\cite{kong2023fftformer} & CVPR'23 & 16.60 & 25.409 & 0.7571 & 15.285 & 42,289 & 23.019 & 0.7394 & 25.979 & 57,091 \\
MixNet~\cite{wu2024mixnet} & Neurocomputing'25&  5.22 & 29.430 & 0.8550 & \best{0.405} & 11,235 & 26.910 & 0.8345 & \best{0.680} & 15,167 \\
UHDformer~\cite{wang2024uhdformer} & AAAI'24 &  0.44 & 28.821 & 0.2370 & 6.334 & 985 & 26.481 & 0.2210 & 10.777 & 1,330 \\
MambaIR~\cite{guo2024mambair} & ECCV'24 &  2.01 & \underline{\second{30.400}} & \underline{\second{0.8694}} & 32.027 & 82,666 & \underline{\second{27.850}} & \underline{\second{0.8493}} & 54.419 & 111,599 \\
TriFormer~\cite{ma2025triformer} & ICASSP'25& 1.09 & 28.910 & 0.8450 & 8.035 & 11,283 & 26.440 & 0.8262 & 6.753 & 15,249 \\
AdaIR~\cite{cui2024adair} & ICLR'25 &  1.05 & 28.838 & 0.8516 & 6.872 & 51,895 & 26.338 & 0.8324 & 11.673 & 70,058 \\
D2Net~\cite{wu2024d2net} & WACV'25 &  5.22 & 28.460 & 0.8520 & 7.743 & 11,523 & 26.020 & 0.8337 & 5.952 & 15,556 \\
UHDDIP~\cite{wang2024uhddip} & TCSVT'25&  0.81 & 29.517 & 0.8585 & 1.877 & \underline{\second{711}} & 26.987 & 0.8385 & 3.163 & \underline{\second{960}} \\
ERR~\cite{zhao2025err} & CVPR'25 &  1.13 & 29.720 & 0.8610 & 2.743 & 807 & 27.230 & 0.8420 & 4.397 & 1,089 \\
UHDPromer~\cite{wang2026uhdpromer} & IJCV2026 & 0.74 & 29.527 & 0.8585 & 1.462 & \best{640} & 27.039 & 0.8387 & 1.920 & \best{864} \\[0.5ex]
ARF-IC & Ours & 3.83 & \best{30.838} & \best{0.8816} & 0.725 & 2,799 & \best{28.328} & \best{0.8619} & \underline{\second{0.984}} & 3,571 \\
\bottomrule
\end{tabular}
\end{adjustbox}
\end{table}

\subsubsection{Non-UHD Image Deblurring}
\begin{figure}[t]
    \centering
    \includegraphics[width=\columnwidth]{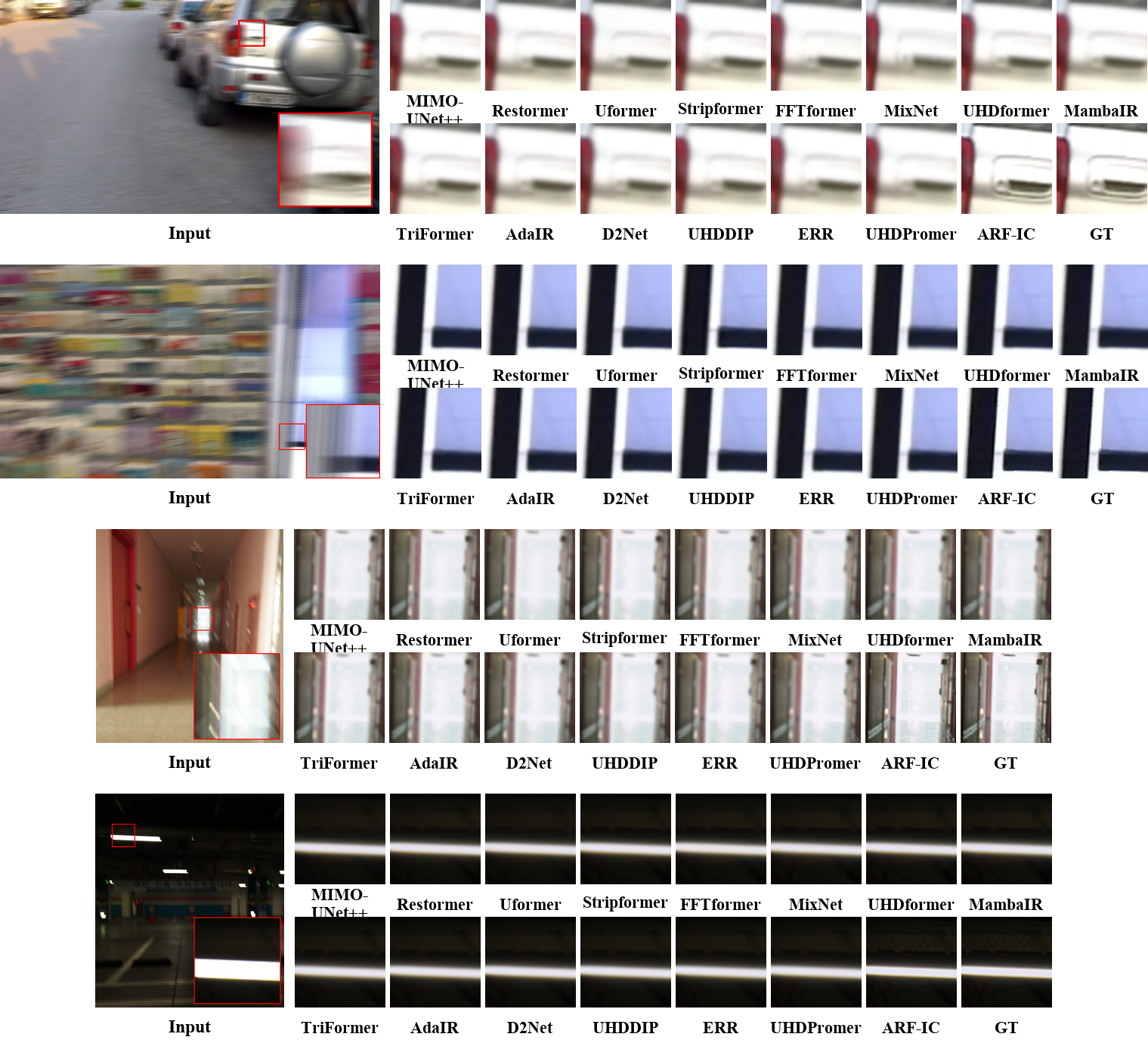}
    \caption{Qualitative comparison on non-UHD deblurring benchmarks. From top to bottom, the rows correspond to GoPro, DVD, RealBlur-R, and RealBlur-J.}
    \vspace{-6mm}
    \label{fig:vis_nonuhd}
\end{figure}

Table~\ref{tab:nonuhd_results} summarizes the quantitative results of ARF-IC on four non-UHD deblurring benchmarks, including per-image inference time (s) and FLOPs (in G). Although our method is primarily designed for UHD image deblurring, it still demonstrates strong generalization and competitive performance on standard-resolution settings. In addition to restoration quality, ARF-IC maintains practical per-image inference time on all four non-UHD benchmarks under our setup.
Qualitative visual comparisons on the four non-UHD benchmarks are shown in Fig.~\ref{fig:vis_nonuhd}.

\begin{table}[t]
\centering
\small
\caption{Quantitative comparison on non-UHD deblurring benchmarks (PSNR, SSIM, per-image inference time in seconds, and FLOPs in G).}
\label{tab:nonuhd_results}
\setlength{\tabcolsep}{2.2pt}

\begin{adjustbox}{width=\columnwidth}
\begin{tabular}{@{}l c c c c c c c c c c c c c c c c@{}}
\toprule
\multirow{2}{*}{Method}
& \multicolumn{4}{c}{GoPro}
& \multicolumn{4}{c}{DVD}
& \multicolumn{4}{c}{RealBlur-R}
& \multicolumn{4}{c}{RealBlur-J} \\
\cline{2-5}\cline{6-9}\cline{10-13}\cline{14-17}
& {PSNR$\uparrow$} & {SSIM$\uparrow$} & {Time$\downarrow$ (s)} & {FLOPs$\downarrow$ (G)}
& {PSNR$\uparrow$} & {SSIM$\uparrow$} & {Time$\downarrow$ (s)} & {FLOPs$\downarrow$ (G)}
& {PSNR$\uparrow$} & {SSIM$\uparrow$} & {Time$\downarrow$ (s)} & {FLOPs$\downarrow$ (G)}
& {PSNR$\uparrow$} & {SSIM$\uparrow$} & {Time$\downarrow$ (s)} & {FLOPs$\downarrow$ (G)} \\
\midrule
MIMO-UNet++~\cite{cho2021mimo} & 32.60 & 0.9580 & 0.1104 & 3,009 & 32.85 & 0.9180 & 0.0936 & 3,009 & 35.70 & 0.9490 & 0.0660 & 2,006 & 27.80 & 0.8420 & 0.0935 & 3,009 \\
Restormer~\cite{zamir2022restormer} & 32.92 & 0.9610 & 0.8462 & 7,454 & 33.52 & 0.9410 & 0.8623 & 7,454 & 36.19 & 0.9570 & 0.5647 & 4,969 & 28.96 & 0.8790 & 0.8464 & 7,454 \\
Uformer~\cite{wang2022uformer}
& 32.97 & \multicolumn{1}{c}{0.9670} & 0.6005 & 4,115
& 33.57 & \multicolumn{1}{c}{0.9470} & 0.6218 & 4,115
& 36.22 & 0.9570 & 0.4101 & 2,744
& \multicolumn{1}{c}{29.06} & \multicolumn{1}{c}{0.8840} & 0.6383 & 4,115 \\
Stripformer~\cite{tsai2022stripformer} & 33.08 & 0.9620 & 0.4510 & 7,409 & 33.68 & 0.9420 & 0.4504 & 7,409 & 36.08 & 0.9540 & 0.3030 & 4,939 & 28.82 & 0.8760 & 0.4511 & 7,409 \\
FFTformer~\cite{kong2023fftformer}
& 34.21 & 0.9680 & 2.2933 & 6,343
& 35.21 & 0.9580 & 2.2938 & 6,343
& 40.11 & 0.9753 & 1.5272 & 4,229
& 32.62 & 0.9326 & 2.2935 & 6,343 \\
MixNet~\cite{wu2024mixnet} & 33.60 & 0.9650 & 0.0592 & 1,685 & 34.05 & 0.9350 & 0.0653 & 1,685 & 36.00 & 0.9560 & 0.0547 & 1,123 & 28.50 & 0.8720 & 0.0735 & 1,685 \\
UHDformer~\cite{wang2024uhdformer} & 31.60 & 0.9500 & 0.3079 & 148 & 31.85 & 0.9100 & 0.3157 & 148 & 34.70 & 0.9420 & 0.2137 & 99 & 27.20 & 0.8250 & 0.3642 & 148 \\
MambaIR~\cite{guo2024mambair} & 34.00 & 0.9660 & 4.7920 & 12,400 & 34.60 & 0.9460 & 4.7858 & 12,400 & 36.20 & 0.9570 & 3.1938 & 8,267 & 28.90 & 0.8780 & 4.7893 & 12,400 \\
TriFormer~\cite{ma2025triformer} & 33.40 & 0.9640 & 0.4580 & 1,693 & 33.85 & 0.9340 & 0.4350 & 1,693 & 35.90 & 0.9530 & 0.4850 & 1,128 & 28.40 & 0.8700 & 0.4940 & 1,693 \\
AdaIR~\cite{cui2024adair}
& \multicolumn{1}{c}{34.20} & \multicolumn{1}{c}{0.9670} & 1.2418 & 7,784
& \multicolumn{1}{c}{34.80} & \multicolumn{1}{c}{0.9470} & 1.2081 & 7,784
& \multicolumn{1}{c}{36.28} & \multicolumn{1}{c}{0.9580} & 0.8448 & 5,189
& 29.00 & 0.8820 & 1.2906 & 7,784 \\
D2Net~\cite{wu2024d2net} & 33.60 & 0.9650 & 0.5790 & 1,728 & 34.05 & 0.9350 & 0.3890 & 1,728 & 36.00 & 0.9560 & 0.3520 & 1,152 & 28.60 & 0.8740 & 0.4520 & 1,728 \\
UHDDIP~\cite{wang2024uhddip} & 32.80 & 0.9600 & 0.2770 & 107 & 33.05 & 0.9200 & 0.2699 & 107 & 35.50 & 0.9490 & 0.1812 & 71 & 28.00 & 0.8600 & 0.2813 & 107 \\
ERR~\cite{zhao2025err} & 33.00 & 0.9610 & 0.4128 & 121 & 33.25 & 0.9210 & 0.3680 & 121 & 35.60 & 0.9500 & 0.2384 & 81 & 28.10 & 0.8620 & 0.3784 & 121 \\
UHDPromer~\cite{wang2026uhdpromer}& 32.79 & 0.9583 & 0.2190 & 96 & 33.03 & 0.9403 & 0.2169 & 96 & 35.27 & 0.9471 & 0.1496 & 64 & 27.92 & 0.8595 & 0.2221 & 96 \\[0.5ex]
ARF-IC (Ours)
& 33.85 & 0.9664 & 0.3021 & 378
& 34.39 & 0.9443 & 0.3311 & 378
& 36.18 & 0.9548 & 0.6534 & 208
& 28.82 & 0.8729 & 0.2948 & 378 \\
\bottomrule
\end{tabular}
\vspace{-4mm}
\end{adjustbox}
\end{table}

\subsection{Ablation Study}

\subsubsection{Experimental Setup}
We conduct ablation studies on the UHD Blur dataset, a UHD synthetic motion blur benchmark with resolutions around 3840 by 2160 and above, containing approximately 2K paired samples. We report PSNR for restoration quality, and we additionally report per-image inference time for settings that affect deployment cost.

\begin{figure*}[t]
\centering
\begin{minipage}[b]{0.32\textwidth}
  \centering
  \includegraphics[width=\linewidth]{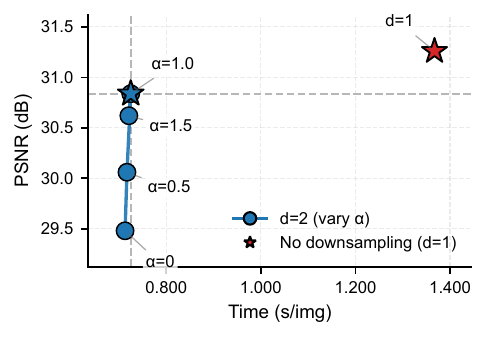}
  {\small (a) Downsampling \& detail weight.}
\end{minipage}\hfill
\begin{minipage}[b]{0.32\textwidth}
  \centering
  \includegraphics[width=\linewidth]{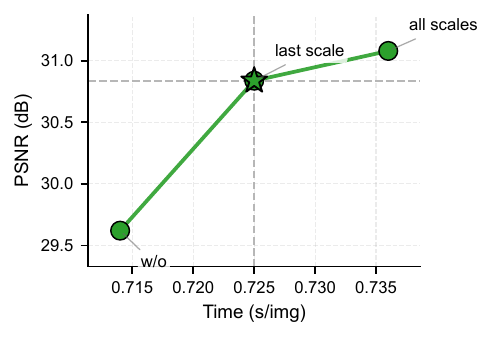}
  {\small (b) Condition regularization.}
\end{minipage}\hfill
\begin{minipage}[b]{0.32\textwidth}
  \centering
  \includegraphics[width=\linewidth]{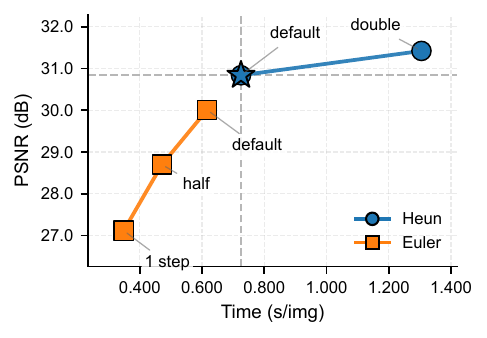}
  {\small (c) ODE solver \& steps.}
\end{minipage}

\vspace{2pt}
\begin{minipage}[b]{0.32\textwidth}
  \centering
  \includegraphics[width=\linewidth]{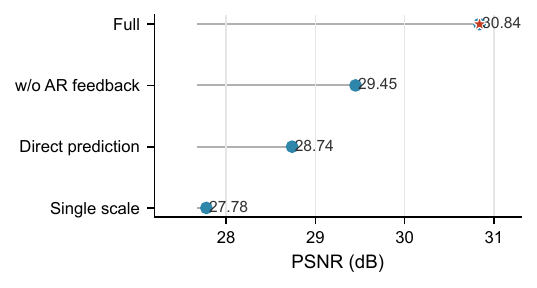}
  {\small (d) MS-ARF core design.}
\end{minipage}\hfill
\begin{minipage}[b]{0.32\textwidth}
  \centering
  \includegraphics[width=\linewidth]{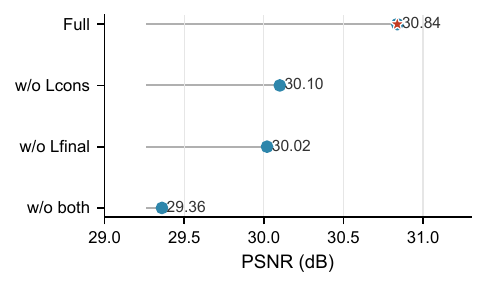}
  {\small (e) Loss terms.}
\end{minipage}\hfill
\begin{minipage}[b]{0.32\textwidth}
  \centering
  \includegraphics[width=\linewidth]{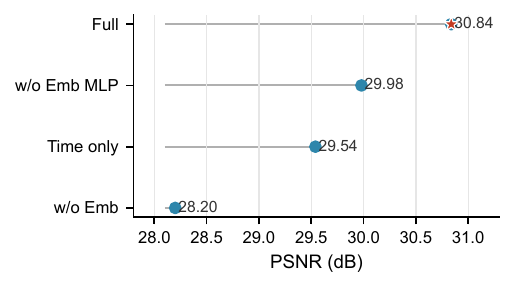}
  {\small (f) Time--scale embedding.}
\end{minipage}

\caption{Ablation study visualizations. Top row: PSNR--time trade-offs for settings affecting runtime. Bottom row: PSNR-only ablations.}
\vspace{-4mm}
\label{fig:ablation_study}
\end{figure*}

\subsubsection{Autoregressive residual modeling}
MS ARF performs autoregressive fusion over a scale sequence by propagating an upsampled previous estimate to the current scale and adding a predicted residual. We ablate by removing the autoregressive feedback input, removing the multi-scale progression, and removing the residual formulation by predicting sharp images directly. Fig.~\ref{fig:ablation_study}(d) shows that removing autoregressive feedback reduces PSNR to 29.45 dB, while removing multi-scale progression leads to a larger drop to PSNR 27.78 dB. 

\subsubsection{Resolution aware downsampling}
To accelerate our method inference, we downsample the input by a factor of 2 $\times$, solve in low resolution, and then upsample back. To compensate for high frequency loss, we show an analytical approach as the difference between the original blurry image and its downsampled and then upsampled reconstruction, and inject it with a detail weight (see Fig.~\ref{fig:ablation_study}(a)). We also attempted to use adaptive pooling on the input images, but the results were unsatisfactory.
%


\subsubsection{Condition number regularization}
To mitigate ill conditioning in UHD scenarios, we introduce condition number regularization on a feature-induced attention matrix, and ablate removing it or applying it at all scales. Fig.~\ref{fig:ablation_study}(b) shows that removing this regularization decreases PSNR to 29.62 dB, suggesting reduced stability in UHD settings. Applying the regularization at all scales further improves quality but increases inference time, indicating that stronger stability constraints may introduce additional computation.

\subsubsection{ODE solver and sampling steps}
At inference, we solve an ODE using numerical integration. We use Heun by default, degenerate to Euler when using one step per scale, and adopt a coarse-to-fine step schedule. Fig.~\ref{fig:ablation_study}(c) demonstrates a clear quality-speed trade-off. Euler is faster than Heun under the same step budget but yields lower PSNR. 


\subsubsection{Loss terms and embedding}
We evaluate the contribution of the cheap final supervision term and the consistency constraint term. Fig.~\ref{fig:ablation_study}(e) shows that disabling the consistency term reduces PSNR to 30.10 dB, indicating that trajectory consistency improves endpoint agreement. Removing the final supervision term yields a further drop to PSNR 30.02 dB, suggesting that one-pass supervision complements flow matching by providing a denser training signal. 
. Fig.~\ref{fig:ablation_study}(f) shows that removing scale conditioning and the embedding using concatenation only also degrades results. 

\section{Discussion and Conclusion}
Beyond the aforementioned ablation experiments, we discuss the approach's performance on downstream tasks (MS COCO 2017~\cite{lin2014coco}), across datasets (image deraining dataset Rain100L~\cite{yang2017deep}), and on edge devices (on the iPhone 16 Pro, our method achieves millisecond-level response times for rendering a 1080p blurred image) (see supplementary materials). 

This paper focuses on UHD image deblurring, balancing 4K/8K fine-detail restoration and inference efficiency via an autoregressive flow framework stabilized by ill-conditioning constraints. It decomposes restoration into a coarse-to-fine residual process, using Rectified Flow for few-step ODE sampling with efficient solvers. The condition-number regularization suppresses UHD numerical instability and hallucinated textures, improving convergence. 
Extensive experiments validate the effectiveness of our approach.

\bibliographystyle{splncs04}
\bibliography{main}

\end{document}